\documentclass[runningheads]{llncs}
\usepackage{graphicx}
\usepackage{multirow}
\usepackage{pifont}
\usepackage{marvosym}

\begin{document}
\title{FGFusion: Fine-Grained Lidar-Camera Fusion for 3D Object Detection}
\author{Zixuan Yin\inst{1} \and Han Sun\inst{1}\textsuperscript{\Letter} \and Ningzhong Liu\inst{1} \and Huiyu Zhou\inst{2} \and Jiaquan Shen\inst{3}}
\authorrunning{Z. Yin et al.}
\institute{Nanjing University of Aeronautics and
Astronautics, Nanjing, China 
\email{sunhan@nuaa.edu.cn}
\and University of Leicester, UK \and Luoyang Normal University, Luoyang, China}

\maketitle              % typeset the header of the contribution
\begin{abstract}
Lidars and cameras are critical sensors that provide complementary information for 3D detection in autonomous driving. While most prevalent methods progressively downscale the 3D point clouds and camera images and then fuse the high-level features, the downscaled features inevitably lose low-level detailed information. In this paper, we propose Fine-Grained Lidar-Camera Fusion (FGFusion) that make full use of multi-scale features of image and point cloud and fuse them in a fine-grained way. First, we design a dual pathway hierarchy structure to extract both high-level semantic and low-level detailed features of the image. Second, an auxiliary network  is introduced to guide point cloud features to better learn the fine-grained spatial information. Finally, we propose multi-scale fusion (MSF) to fuse the last N feature maps of image and point cloud. Extensive experiments on two popular autonomous driving benchmarks, i.e. KITTI and Waymo, demonstrate the effectiveness of our method.

\end{abstract}

\section{Introduction}
3D object detection is a crucial task in autonomous driving\cite{arnold2019survey,huang2022multi}. In recent years, lidar-only methods have made significant progress in this field. However, relying solely on point cloud data is insufficient because lidar only provides low-resolution shape and depth information. Therefore, researchers hope to leverage multiple modalities of data to improve detection accuracy. Among them, vehicle-mounted cameras can provide high-resolution shape and texture information, which is complementary to lidar. Therefore, the fusion of point cloud data with RGB images has become a research hotspot.

In the early stages of fusion method research, researchers naturally assumed that the performance of fusion methods would be better than that of lidar-only methods, because the essence of fusion methods is to add RGB information as an auxiliary to lidar-only methods. Therefore, the performance of the model should be at least as good as before, rather than declining\cite{vora2020pointpainting}. However, this is not always the case.

There are two reasons for the performance decline: 1) a suitable method for aligning the two modal data has not yet been found, 2) the features of the two modalities used in the fusion are too coarse. Regarding the first issue, fusion methods have evolved from the initial post-fusion\cite{chen2017multi,ku2018joint} and point-level fusion\cite{vora2020pointpainting,wang2021pointaugmenting} methods to today’s more advanced feature fusion\cite{bai2022transfusion,li2022deepfusion,wu2022sparse} methods. However, the second problem has not yet been solved. Specifically, we know that lidar-only methods are mainly divided into one-stage methods\cite{zhou2018voxelnet,liu2020tanet,yang20203dssd} and two-stage methods\cite{chen2019fast,shi2020pv,shi2019pointrcnn,shi2020points}. Usually, the performance of two-stage methods is better than that of one-stage methods because the features extracted by the first stage can be refined in the second stage. However, most current fusion methods focus on how to fuse features more effectively and ignore the process of refining fused features.

To solve the above problems, we utilize fine-grained features to improve the model accuracy and propose an efficient multi-modal fusion strategy called FGFusion. Specifically, since both image and point cloud data inevitably lose detailed features and spatial information during the downscaling process, we design different feature refinement schemes for the two modalities. First, for image data, we exploit a dual-path pyramid structure and designs a top-down feature path and a bottom-up attention path to better fuse high-level and low-level features. For point cloud data, inspired by SASSD\cite{he2020structure}, we construct an auxiliary network with point-level supervision to guide the intermediate features from different stages of 3D backbone to learn the fine-grained spatial structures of point clouds. In the fusion stage, we select several feature maps of the same number from the feature pyramids of images and point clouds respectively, and fuse them by cross-attention. The fused feature pyramids can then be passed into modern task prediction head architecture\cite{yin2021center,bai2022transfusion}.

In brief, our contributions can be summarized as follows:
\begin{itemize}
    \item We design different feature refinement schemes for camera image and point cloud data, in order to fuse high-level abstract semantic information and low-level detailed features.
    \item We design a multi-level fusion strategy for point clouds and images, which fully utilizes the feature pyramids of the two modalities in the fusion stage to improve the model accuracy.
    \item We verify our method on two mainstream autonomous driving point cloud datasets (KITTI and Waymo), and the experimental results prove the effectiveness of our method.
\end{itemize}

\section{Related Work}

\subsection{LiDAR-only 3D Detection}
Lidar-only methods are mainly divided into point-based methods and voxel-based methods. Among them, point-based methods such as PointNet\cite{qi2017pointnet} and PointNet++\cite{qi2017pointnet++} are the earliest neural networks directly applied to point clouds. They directly process unordered raw point clouds and extract local features through max-pooling. Based on their work, voxel-based and pillar-based methods have been derived. They transform the original point cloud into a Euclidean feature space and then use standard 2D or 3D convolution to calculate the features of the BEV plane. Representative methods include VoxelNet\cite{zhou2018voxelnet}, SECOND\cite{yan2018second}, PointPillars\cite{lang2019pointpillars}, etc.

The development of lidar-only methods later shows two different development trends. Like 2D object detection, they are divided into one-stage and two-stage methods. One-stage methods\cite{zhou2018voxelnet,liu2020tanet,yang20203dssd} directly regress category scores and bounding boxes in one stage, and the network is relatively simple and has fast inference speed. Two-stage methods\cite{chen2019fast,shi2020pv,shi2019pointrcnn,shi2020points} usually generate region proposals in the first stage and then refine them in the second stage. The accuracy of two-stage methods is usually higher than that of one-stage methods, because the second stage can capture more detailed and distinctive features, but the cost is a more complex network structure and higher computational cost.

\subsection{Fusion-based 3D Detection}
Due to the sparsity of point cloud data and its sole possession of spatial structural information, researchers have proposed to complement point clouds with RGB images. Early methods\cite{chen2017multi,qi2018frustum} use result-level or proposal-level post-fusion strategies, but the fusion granularity is too coarse, resulting in performance inferior to that of lidar-only methods.

PointPainting\cite{vora2020pointpainting} is the first to utilize the hard correlation between LiDAR points and image pixels for fusion. It projects the point clouds onto the images through a calibration matrix and enhances each LiDAR point with the semantic segmentation score of the image. PointAugmenting\cite{wang2021pointaugmenting} builds on PointPainting and proposes using features extracted from 2D object detection networks instead of semantic segmentation scores to enhance LiDAR points. Feature-level fusion methods points out that the hard association between points and pixels established by the calibration matrix is unreliable. DeepFusion\cite{li2022deepfusion} uses cross-attention to fuse point cloud features and image features. TransFusion\cite{bai2022transfusion} uses the prediction of point cloud features as a query for image features and then uses a transformer-like architecture to fuse features.

It can be seen that whether using semantic segmentation scores and image features obtained from pre-trained networks, or directly querying and fusing at the feature level, these methods essentially fuse high-level features with the richest semantic information, while ignoring low-level detailed information.

\section{FGFusion}
\subsection{Motivations and Pipeline}
\begin{figure}
\includegraphics[width=\textwidth]{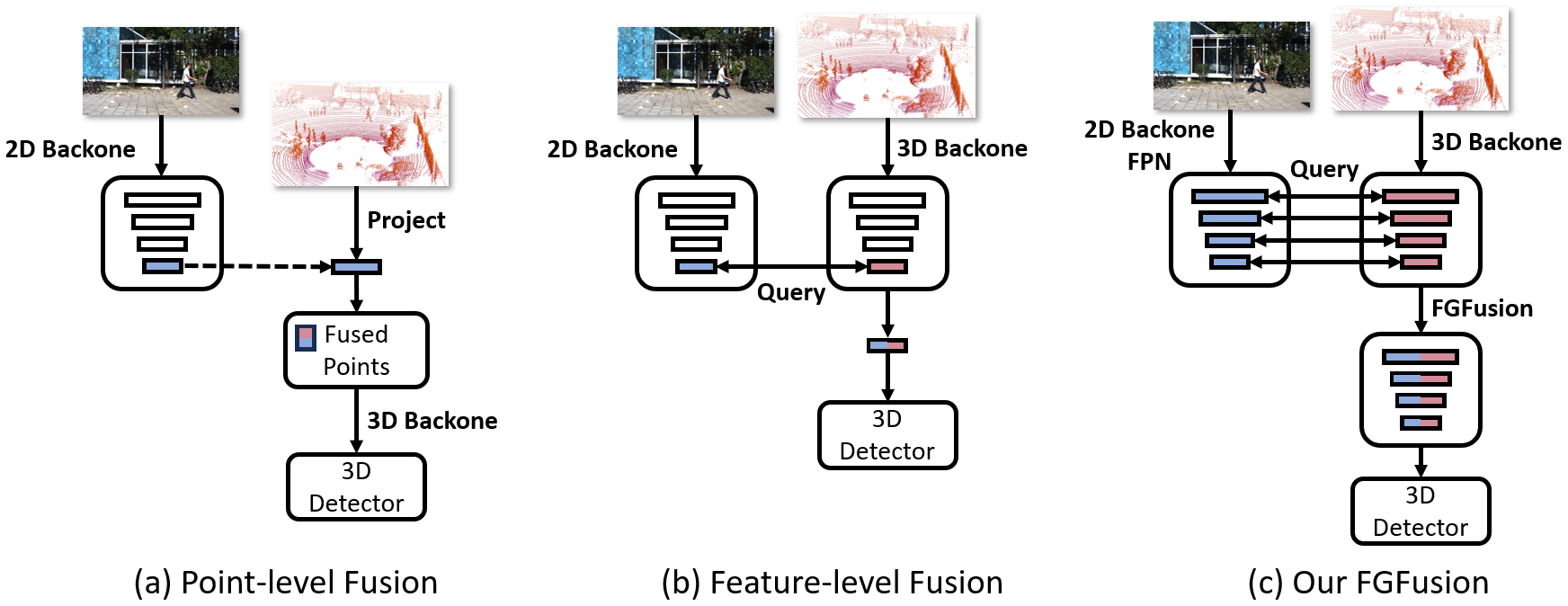}
\caption{Most point-level fusion methods\cite{vora2020pointpainting,wang2021pointaugmenting} and feature-level fusion methods\cite{li2022deepfusion,bai2022transfusion} only use the last layer of image or point cloud features for fusion, while our FGFusion performs fusion at multiple feature scales, fully utilizing low-level detail information to improve model accuracy.} \label{fig_compare}
\end{figure}

The previous fusion methods only exploit high-level features, ignoring the important fact that detailed feature representations are lost in the downsampling process. For example, PointPainting\cite{vora2020pointpainting} directly makes use of pixel-wise semantic segmentation scores as image features to decorate point cloud data, which only uses the results of last feature map and ignores multi-scale information. PointAugmenting\cite{wang2021pointaugmenting} utilizes the last feature map with the richest semantic information to decorate point cloud data, but discards all the others that contain low-level detailed information. DeepFusion\cite{li2022deepfusion} is a feature-level fusion method, which improves the accuracy compared to point-level fusion methods such as PointAugmenting, but the essence is the same, as shown in Fig.~\ref{fig_compare}.

We noticed that in some 2D object detection tasks, such as small object detection and fine-grained image recognition, multi-scale techniques are often used to extract fine-grained features. While in 3D object detection, point cloud data is suitable for capturing the spatial structural features, but it is easy to ignore small targets and fine features due to its sparse characteristic. Therefore, we hope to fuse point cloud and image data in a multi-scale way to make up for the shortcomings of point clouds. To achieve this goal, we fuse the features of point cloud and image at multiple levels instead of only using the last feature map generated by the backbone network. In addition, to extract finer features, we design a dual-path pyramid structure in the image branch and add an auxiliary network to guide convolutional feature perception of object structures in the point cloud branch.

\begin{figure}
\includegraphics[width=\textwidth]{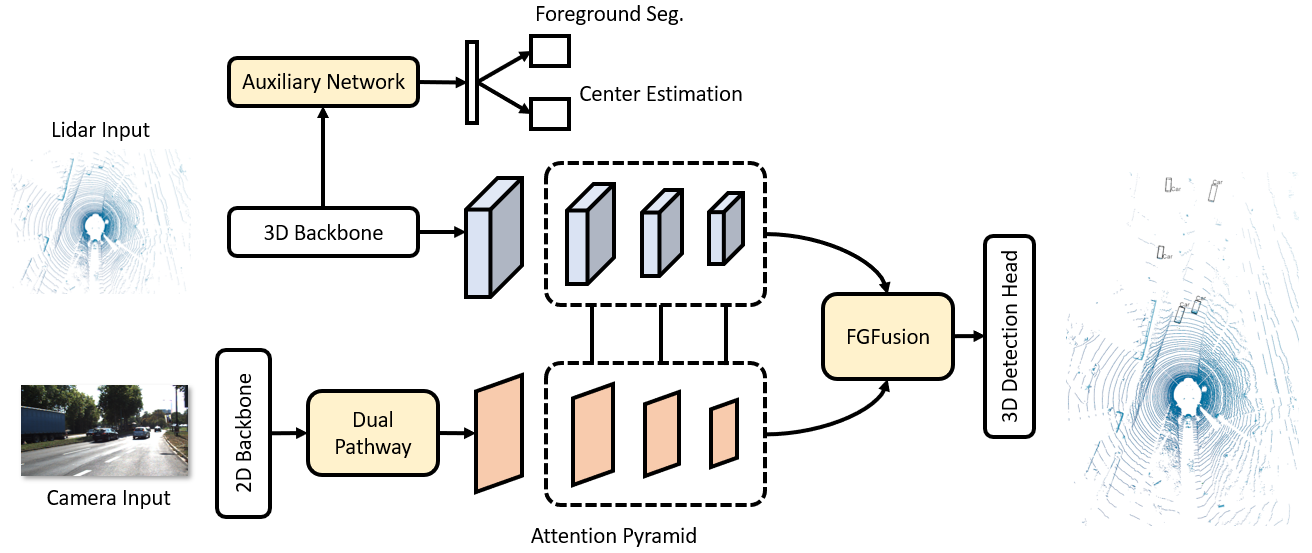}
\caption{An overview of FGFusion framework. FGFusion consists of 1) a dual pathway hierarchy structure with a top-down feature pathway and a bottom-up attention pathway, hence learning both high-level semantic and low-level detailed feature representation of the images, 2) an auxiliary network to guide point cloud features to better learn the fine-grained spatial information, and 3) a fusion strategy that can fuse the two modalities in a multi-scale way.} \label{fig_pipeline}
\end{figure}

To summarize, our proposed fine-grained fusion pipeline is shown in Fig.~\ref{fig_pipeline}. For the image branch, we exploit 2D backbone and a dual-path pyramid structure to obtain the attention pyramid. For the point cloud branch, the raw points are fed into the existing 3D backbone to obtain the lidar features, and at the same time, guide the learning of features through an auxiliary network. Finally, we fuse the image and point cloud features at different levels and attach the same designed head to each fused layer of features to obtain the final results.

\subsection{Camera Stream Architecture}

In general, the input image will be processed by a convolutional neural network to obtain a feature representation with high-level semantic information. However, many low-level detailed features will be lost, which is insufficient for robust fusion. In order to retain the fine-grained features, inspired by the FPN network\cite{lin2017feature}, we design a top-down feature path to extract features of different scales.

Let $\{B_1,B_2,...,B_l\}$ represent the feature maps obtained after the input image passes through the backbone and $l$ represent the number of convolutional blocks. The general method is to directly use the output of the last block $B_l$ for fusion, but we hope to make full use of each $B_i$. Since it will bring huge cost overheads inevitably if making full use of every blocks of the network, we only select the last $N$ outputs to generate the corresponding feature pyramid. The final feature pyramid obtained can be denoted as $\{F_{l-N+1},F_{l-N+2},...,F_l\}$.

After obtaining the feature pyramid, we design a bottom-up attention path which includes spatial attention and channel attention. Spatial attention is used to locate the identifiable regions of the input image at different scales. It can be represented as:
\begin{equation}
A_i^s=\sigma(K\ast F_i),
\end{equation}
where $\sigma$ is the sigmoid activation function, $\ast$ represents the deconvolution operation, and $K$ represents the convolution kernel. Channel attention is used to add associations between channels and pass low-level detailed information layer by layer to higher levels:
\begin{equation}
A_i^c=\sigma (W_b\cdot ReLU(W_a\cdot GAP(F_i))),
\end{equation}
where $\cdot$ represents element-wise multiplication, $W_a$ and $W_b$ represent the weight parameters of two fully connected layers. $\rm{GAP}(\cdot)$ represents global average pooling. In order to transmit low-level detailed information to high-level features, $A_i^c$ need to be added with $A_{i-1}^c$ and then downsampled twice to generate a bottom-up path.

After obtaining the attention pyramid, a bottom-up attention path can be generated in combination with the spatial pyramid. Specifically, this paper first adds spatial attention $A_i^s$ and channel attention $A_i^c$, and then performs dot product operation with $F_i$ in the feature pyramid to obtain $F_i'$:
\begin{equation}
F_i'=F_i\cdot(A_i^s+\alpha A_i^c).
\end{equation}
Finally, $\{F_{l-N+1}',F_{l-N+2}',...,F_l'\}$ can be obtained for subsequent classification.

\subsection{LiDAR Stream Architecture}
Our framework can use any network that can convert point clouds into multi-scale feature pyramids as our lidar flow. At the same time, inspired by SASSD\cite{he2020structure}, we designed an auxiliary network, which contains a point-wise foreground segmentation head and a center estimation head, to guide the backbone CNN to learn the fine-grained structure of point clouds at different stages of intermediate feature learning. It is worth noting that the auxiliary network can be separated after training, so no additional computation is introduced during inference.

\begin{figure}
\includegraphics[width=\textwidth]{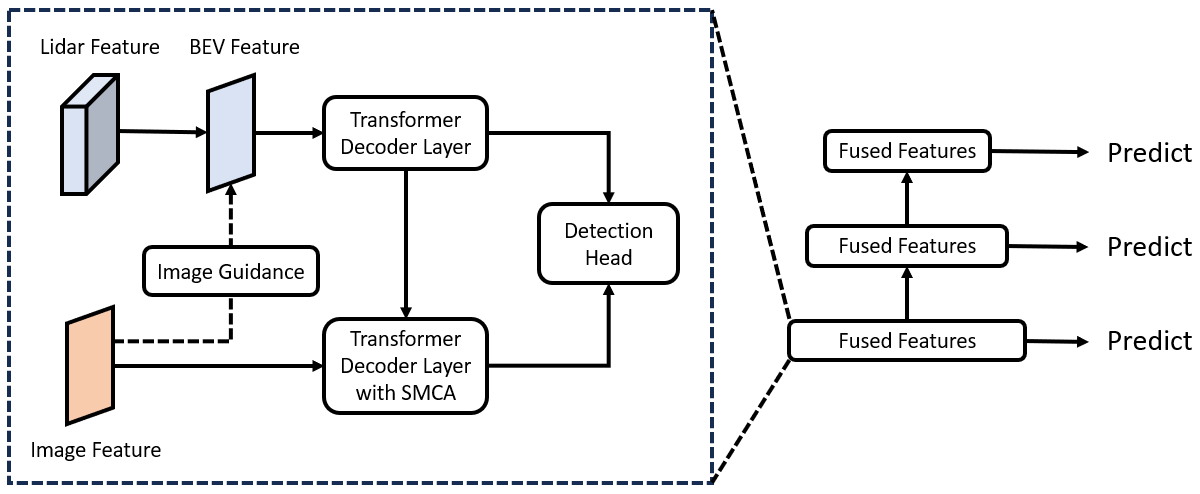}
\caption{The multi-scale fusion module first compresses the point cloud features into BEV features, and then uses TransFusion\cite{bai2022transfusion} to fuse the last N layers of BEV features and image features separately to obtain the prediction results of each layer. Finally, the post-processing is performed to obtain the final results.} \label{fig_MSF}
\end{figure}

\subsection{Multi-scale Fusion Module}

Now we have obtained the attention pyramid of the image and the feature pyramid of the point cloud separately. In order to fully fuse the two modalities, we take the last $N$ layers of features of both for fusion, rather than just using the last layer, as shown in Fig.~\ref{fig_MSF}. Through the point cloud feature pyramid, we can obtain a multi-scale point cloud BEV feature map $\{F^B_{l-N+1},F^B_{l-N+2}...,F^B_{l}\}$. Following TransFusion\cite{bai2022transfusion}, we use two transformer decoding layers to fuse the two modalities: first decodes object queries into initial bounding box predictions using the LiDAR information, and then performs LiDAR-camera fusion by attentively fusing object queries with useful image features. Finally, each fusion feature can generate corresponding prediction results, and the final prediction is obtained through post-processing.

\section{Experiments}
We evaluate our proposed FGFuison on two datasets, KITTI\cite{geiger2012we} and Waymo\cite{sun2020scalability}, and conduct sufficient ablation experiments.

\subsection{Datasets}
The KITTI dataset contains 7481 training samples and 7518 testing samples of autonomous driving scenes. As common practice, we divide the training data into a training set containing 3712 samples and a validation set containing 3769 samples. According to the requirements of the KITTI object detection benchmark, we conduct experiments on three categories of cars, pedestrians, and cyclists and evaluate the results using the average precision (AP) with an IoU threshold of 0.7.

The Waymo Open Dataset contains 798 training sequences, 202 validation sequences and 150 testing sequences. Each sequence has about 200 frames, which contain lidar points, camera images, and labeled 3D bounding boxes. We use official metrics, i.e., Average Precision (AP) and Average Precision weighted by Heading (APH), to evaluate the performance of different models and report the results of LEVEL1 (L1) and LEVEL2 (L2) difficulty levels.

\subsection{Implementation Details}
For the KITTI dataset, the voxel size is set to (0.05m, 0.05m, 0.1m). Since KITTI only provides annotations for the front camera’s field of view, the detection range of the X, Y and Z axes are set to [0, 70.4 m], [-40m, 40m], and [-3m, 1m], respectively. The image size is set to 448 × 800. For the Waymo dataset, the voxel size is set to (0.1m, 0.1m, 0.15m). The detection range of the X and Y axes is [-75.2m, 75.2m], and the detection range of the Z axis is [-2m, 4m]. 

We choose TransFusion-L and the DLA34 of the pre-trained CenterNet as the 3D and 2D backbone networks, respectively. Following TransFusion\cite{bai2022transfusion}, our training consists of two stages: 1) First we train the 3D backbone with the first decoder layer and FFN for 20 epochs. It only requires point clouds as input, and the last BEV feature map is used to produce initial 3D bounding box predictions. 2) Then we train the LiDAR-camera fusion and image-guided query initialization module for another 6 epochs. In this stage, the last three feature maps of the 3D and 2D backbone are fused separately. The advantage of this two-step training scheme over joint training is that auxiliary networks can be used only in the first stage, as well as data augmentation methods for pure point cloud methods. For post-processing, we use NMS with the threshold of 0.7 for Waymo and 0.55 for KITTI to remove redundant boxes.

\subsection{Experimental Results and Analysis}
\begin{table}
\caption{Performance comparison on the KITTI \emph{val} set with AP calculated by 40 recall positions.}
\centering
\begin{tabular}{c|c|c|c|c|c|c|c|c|c|c|c}
\hline
\multirow{2}*{Method} & \multirow{2}*{Modality} & \multirow{2}*{mAP} & \multicolumn{3}{c|}{Car} & \multicolumn{3}{c|}{Pedestrian} & \multicolumn{3}{c}{Cyclist}\\
\cline{4-12}
& & & Easy & Mod. & Hard & Easy & Mod. & Hard & Easy & Mod. & Hard\\
\hline %[2pt]
SECOND\cite{yan2018second} & L & 68.06 & 88.61 & 78.62 & 77.22 & 56.55 & 52.98 & 47.73 & 80.58 & 67.15 & 63.10\\
PointPillars\cite{lang2019pointpillars} & L & 66.53 & 86.46 & 77.28 & 74.65 & 57.75 & 52.29 & 47.90 & 80.05 & 62.68 & 59.70\\
PointRCNN\cite{shi2019pointrcnn} & L & 70.67 & 88.72 & 78.61 & 77.82 & 62.72 & 53.85 & 50.25 & 86.84 & 71.62 & 65.59\\
PV-RCNN\cite{shi2020pv} & L & 73.27 & 92.10 & 84.36 & 82.48 & 64.26 & 56.67 & 51.91 & 88.88 & 71.95 & 66.78\\
Voxel-RCNN\cite{deng2021voxel} & L & - & \textbf{92.38} & \textbf{85.29} & 82.86 & - & - & - & - & - & -\\
\hline
MV3D\cite{chen2017multi} & L+C & - & 71.29 & 62.68 & 56.56 & - & - & - & - & - & -\\
AVOD\cite{ku2018joint} & L+C & - & 84.41 & 74.44 & 68.65 & - & 58.80 & - & - & 49.70 & -\\
F-PointNet\cite{qi2018frustum} & L+C & 65.58 & 83.76 & 70.92 & 63.65 & 70.00 & 61.32 & 53.59 & 77.15 & 56.49 & 53.37\\
3D-CVF\cite{yoo20203d} & L+C & - & 89.67 & 79.88 & 78.47 & - & - & - & - & - & -\\
EPNet\cite{huang2020epnet} & L+C & 70.97 & 88.76 & 78.65 & 78.32 & 66.74 & 59.29 & 54.82 & 83.88 & 65.60 & 62.70\\
CAT-Det\cite{zhang2022cat} & L+C & 75.42 & 90.12 & 81.46 & 79.15 & \textbf{74.08} & \textbf{66.35} & 58.92 & 87.64 & 72.82 & 68.20\\
\hline
FGFusion(Ours) & L+C & \textbf{77.05} & \textbf{92.38} & 84.96 & \textbf{83.84} & 72.63 & 65.07 & \textbf{59.21} & \textbf{90.33} & \textbf{74.19} & \textbf{70.84}\\
\hline
\end{tabular}
\end{table}

\subsubsection{KITTI.}
To prove the effectiveness of our method, we compare the average precision (AP) of FGFusion with some state-of-the-art methods on the KITTI dataset. As shown in Table 1, the mAP of our proposed FGFusion is the highest among all methods. KITTI divides all objects into three difficulty levels: easy, moderate and hard based on the size of the object, occlusion status and truncation level. The higher the difficulty level, the harder it is to detect. Our method leads in different levels of difficulty for multiple categories and has higher accuracy than all other methods in the difficult levels of all three categories, which proves that our method can effectively fuse fine-grained features.

In lidar-only methods, the accuracy of one-stage methods such as SECOND\cite{yan2018second} and PointPillars\cite{lang2019pointpillars} is lower than that of two-stage methods such as PV-RCNN\cite{shi2020pv}. In the easy and medium difficulty levels of the car category, our FGFusion is competitive with Voxel-RCNN\cite{deng2021voxel}, the best-performing method in lidar-only methods, and surpasses 0.98\% AP in the difficult level. In fusion methods, early works such as MV3D\cite{chen2017multi} and AVOD\cite{ku2018joint} have lower performance than lidar-only methods. However, recently proposed CAT-Det\cite{zhang2022cat} can achieve higher overall accuracy than lidar-only methods in all three categories, and achieve 75.42 in mAP, which is a little lower than that of our method.

\begin{table}
\caption{Performance comparison on the Waymo \emph{val} set for 3D vehicle (IoU = 0.7) and pedestrian (IoU = 0.5) detection.}
\centering
\begin{tabular}{c|c|c|c|c|c}
\hline
\multirow{2}*{Method} & \multirow{2}*{Modality} & \multicolumn{2}{c|}{Vehicle(AP/APH)} & \multicolumn{2}{c}{Pedestrian(AP/APH)}\\
\cline{3-6}
    &       & L1 & L2 & L1 & L2\\
\hline %[2pt]
SECOND\cite{yan2018second} & L & 72.27/71.69 & 63.85/63.33 & 68.70/58.18 & 60.72/51.31\\
PointPillars\cite{lang2019pointpillars} & L & 71.60/71.00 &  63.10/62.50 & 70.60/56.70 & 62.90/50.20\\
PV-RCNN\cite{shi2020pv} & L & 77.51/76.89 & 68.98/68.41 & 75.01/65.65 & 66.04/57.61\\
CenterPoint\cite{yin2021center} & L & - & -/66.20 & - & -/62.60\\
3D-MAN\cite{yang20213d} & L & 74.50/74.00 & 67.60/67.10 & 71.70/67.70 & 62.60/59.00\\
PointAugmenting\cite{wang2021pointaugmenting} & L+C & 67.40/- & 62.70/- & 75.04/- & 70.60/-\\
DeepFusion\cite{li2022deepfusion} & L+C & 80.60/80.10 & 72.90/72.40 & \textbf{85.80}/\textbf{83.00} & 78.70/76.00\\
\hline
FGFusion(Ours) & L+C & \textbf{81.92}/\textbf{81.44} & \textbf{73.85}/\textbf{73.34} & 85.73/82.85 & \textbf{78.81}/\textbf{76.14}\\
\hline
\end{tabular}
\end{table}

\subsubsection{Waymo.}
Compared with the KITTI dataset, the Waymo dataset is larger and more diverse in sample diversity, and hence is more challenging. To verify our proposed FGFusion, we also conduct experiments on the Waymo dataset and compare it with some state-of-the-art methods. Table 2 shows that our FGFusion is better than other methods for both car and pedestrian categories in LEVEL2 difficulty, which is the main metric for ranking in the Waymo 3D detection challenge. Compared with the best PV-RCNN\cite{shi2020pv} in lidar-only methods, FGFusion has improved the APH of vehicle recognition by 4.93\% and that of pedestrian recognition by 18.53\%, which proves that our fusion method is more advantageous in small object detection.

\subsection{Ablation study}

We conduct a series of experiments on Waymo to demonstrate the effectiveness of each component in our proposed FGFusion, including the attention pyramid of the image branch (AP), the auxiliary network of the point cloud branch (AN), and the multi-scale fusion module (MSF).

\begin{table}
\caption{Effect of each component in FGFusion on Waymo \emph{val} set with APH in L2 difficulty. }
\centering
\begin{tabular}{ccc|cc}
\hline
MSF & AP & AN & Vehicle & Pedestrian\\
\hline %[2pt]
& & & 70.42 & 72.94\\
\ding{51} & & & 72.16 & 74.87\\
\ding{51} & \ding{51} & & 72.86 & 75.74\\
\ding{51} & & \ding{51} & 72.77 & 75.38\\
\ding{51} & \ding{51} & \ding{51} & 73.34 & 76.14\\
\hline
\end{tabular}
\end{table}

\begin{table}
\caption{Performance comparison on Waymo \emph{val} set with APH in L2 difficulty using different number of features for fusion.}
\centering
\begin{tabular}{c|cc}
\hline
Feature Num. & Vehicle & Pedestrian\\
\hline %[2pt]
1 & 70.42 & 72.94\\
2 & 71.62 (+1.20) & 74.05 (+1.11)\\
3 & 72.16 (+0.54) & 74.81 (+0.76)\\
4 & 72.32 (+0.16) & 75.01 (+0.20)\\
\hline
\end{tabular}
\end{table}

\subsubsection{Effect of each component.}
As shown in Table 3, our FGFusion is 2.92\% and 3.2\% higher than the baseline in APH for the two categories, vehicles and pedestrians, respectively. Specifically, the multi-scale fusion module brings improvements of 1.74\% and 1.93\% to the baseline on two categories, which confirms our proposed fine-grained fusion strategy. The attention pyramid or the auxiliary network can further bring improvements of (0.7\%, 0.87\%) and (0.61\%, 0.51\%), respectively. This indicates that the finer the fused features, the higher the model accuracy can achieve, which is consistent with our expectation.

\subsubsection{Number of feature layers selected for fusion.}

The number of fusion features for point clouds and images is the key hyperparameter of our multi-scale fusion module. In order to determine the optimal value, we conduct experiments on the Waymo dataset without using attention pyramids or auxiliary networks. As shown in Table 4, the more feature layers used, the higher the model accuracy can achieve. This is because high-level features have rich semantic information and low-level features reserve complementary detailed information. The more feature layers used for fusion, the less information lost during downsampling. From the experimental results, it is intuitive that using two or three layers of features for fusion can bring significant improvements to model accuracy. While the number of fusion layers reaches four, the degree of improvement will be greatly reduced. It is worth noting that the more fusion layers used, the more weights the cross-attention model needs to train during fusion. In order to balance between model accuracy and computational cost, we use three layers of features for fusion in our experiments.

\section{Conclusion}
In this paper, we propose a novel multimodal network FGFusion for 3D object detection in autonomous driving scenarios. We design fine-grained feature extraction networks for both the point cloud branch and the image branch, and fuse features from different levels through a pyramid structure to improve detection accuracy. Extensive experiments are conducted on the KITTI and Waymo datasets, and the experimental results show that our method can achieve better performance than some state-of-the-art methods.

%
% ---- Bibliography ----
%
\bibliographystyle{splncs04}
\bibliography{FGFusion}
\end{document}